\pdfoutput=1

\documentclass[11pt]{article}

\usepackage{acl}

\usepackage{amssymb}
\usepackage{amsmath}
\usepackage[export]{adjustbox}
\usepackage{graphicx}
\usepackage{float}
\usepackage{multirow}
\usepackage{tabularx}
\usepackage{booktabs}
\usepackage{arydshln}
\usepackage{hyperref}

\usepackage{times}
\usepackage{latexsym}

\usepackage[T1]{fontenc}

\usepackage[utf8]{inputenc}

\usepackage{microtype}

\usepackage{inconsolata}

\title{IPED: An Implicit Perspective for Relational Triple Extraction based on Diffusion Model}

\author{Jianli Zhao\hspace{1em} \   Changhao Xu\hspace{1em} \   Bin Jiang \\
  Shandong University \\
  \texttt{\{jianliz, xch, jiangbin\}@mail.sdu.edu.cn} \\}

\begin{document}
\maketitle
\begin{abstract}
Relational triple extraction is a fundamental task in the field of information extraction, and a promising framework based on table filling has recently gained attention as a potential baseline for entity relation extraction. However, inherent shortcomings such as redundant information and incomplete triple recognition remain problematic. To address these challenges, we propose an \textbf{I}mplicit \textbf{P}erspective for relational triple \textbf{E}xtraction based on \textbf{D}iffusion model (IPED), an innovative approach for extracting relational triples. Our classifier-free solution adopts an implicit strategy using block coverage to complete the tables, avoiding the limitations of explicit tagging methods. Additionally, we introduce a generative model structure, the block-denoising diffusion model, to collaborate with our implicit perspective and effectively circumvent redundant information disruptions. Experimental results on two popular datasets demonstrate that IPED achieves state-of-the-art performance while gaining superior inference speed and low computational complexity. To support future research, we have made our source code publicly available online. \footnote{Our source code responsity is released at: \href{https://github.com/girlsuuu/IPED}{https://github.com/girlsuuu/IPED}.}
\end{abstract}

\section{Introduction}

The extraction of relational triples has been an important and fundamental task in knowledge graph construction \cite{info,casrel}, aiming to recognize triples in the form of (\textit{head entity}, \textit{relation}, \textit{tail entity}) from unstructured text. Current research in information extraction can be categorized into two main approaches: the joint extraction models, which utilize a simultaneous style, and the pipeline models, which utilize a two-encoder methodology to extract entities and relations. While the pipeline framework is criticized for serious error propagation and lack of interaction between its two subtasks \cite{trigger}, leading to performance decline, many recent joint extraction models have begun to thrive due to their enhanced capability to deal with complex scenarios such as single entity overlap (SEO), entity pair overlap (EPO), and subject object overlap (SOO).

Among these popular joint extraction methods, one baseline, known as the table-filling method, has gained favor in recent research. Compared to a multi-task joint structure, this method features a table of token pair units that are to be filled and decoded in a single step. In this way, it avoids exposure bias and error propagation, challenges that most methods cannot fully overcome. Particularly for recently proposed models \cite{onerel, grte, unire}, these can employ a novel table-filling strategy to simplify the decoding process and enhance information interaction.

Despite many unique advantages over table-filling methods, some flaws still remain to be addressed. (1) The abundance of negative tagging in a table, which is significantly denser than positive tagging, leads to imbalanced labeling and redundant information \cite{unire, odrte}. To the best of our knowledge, this is a universal issue across all table-filling models. This imbalance results in a bias towards negative tagging and heightened computational complexity. (2) Many table-filling strategies fail to extract all scenarios of triples, leading to decreased recall \cite{odrte}. Even in the recent significant work \cite{onerel}, entities consisting of a single token in a triple cannot be properly extracted due to conflicts arising from multiple labels in one element. (3) Once a sentence contains multiple triples, the separate labels of different triples may intersect in a single element, causing confusion in decoding all ground-truth triples. Many models \cite{grte,odrte} employ decoding algorithms that match labels based on the nearest-neighbor principle, which can lead to error associations within a triple. (4) A line of models, not limited to table-filling ones, exhibit poor learning performance on the WebNLG dataset in contrast to the NYT dataset and they attribute it to the vast number of predefined relations in the former dataset \cite{ergm}.

After conducting a detailed observation and analysis of their models, it is observed that all existing table-filling-based methods are consistently constrained by the approach of utilizing a classifier to tag each table element explicitly. Mainly because of this, most of them can hardly escape the challenges mentioned above, despite attempts to introduce innovative labeling strategies and creative decoding algorithms. This constraint necessitates traversing each element of the table, consequently leading to a substantial number of negative samplings. This explicit way of assigning a fixed label to each element can not cope with scenarios when one element requires multiple labels, leading to the inability to recognize all triples and confusion in the regions where triple labels intersect. Additionally, certain decoding strategies, designed in response to this approach, often result in incorrectly matched labels for a triple.

To address the aforementioned issues at a fundamental level, instead of explicitly labeling all the elements, we formulate a fresh perspective to implicitly fill the tables using a block-covered approach. In this method, blocks defined by four edges (up, down, right, left) and one level are refined within a three-dimensional table (multiple two-dimensional tables stacked together). In alignment with this implicit approach, we introduce a generative model designed to recover all blocks within the tables. Specifically, our proposed block-denoising diffusion model (Blk-DDM) can progressively refine the edges and levels of the initialized blocks step by step through a reverse process, ensuring the blocks precisely cover the ground truth triples horizontally, vertically, and deeply. As a result, our model naturally disregards redundant information by leaving the negative spaces alone rather than classifying them. Furthermore, our approach allows for the adequate recognition of all potential triples, as the proposed blocks can overlap implicitly. In contrast to previous decoding algorithms that match explicit labels, our proposed simple but effective Parallel Boundary Emitting Strategy (PBES) for decoding has the capability of extracting all triples accurately, circumventing error association challenges and significantly accelerating inference. Additionally, our denoising diffusion process enables the gradual refinement of specific fine-grained relation types within triples, enhancing performance in large-relation datasets such as WebNLG (demonstrated in Section \ref{Ablation}). Experimental results on two datasets, NYT and WebNLG, demonstrate that our model achieves state-of-the-art performance and exhibits superior efficiency in inference.

\section{Related Works}
\subsection{Joint Extraction Models}

Existing joint extraction models can be roughly sorted into two frameworks. The first framework, based on multi-task learning, utilizes a shared encoder but employs distinct decoders to sequentially predict entities and relations. \cite{r1} proposes an integrated model that extracts entities and relations separately, leveraging shared parameters and mutual interaction. \cite{r2} adopts a model employing shared data representations to mitigate error propagation between tasks. CasRel \cite{r3} treats relations as functions mapping subjects to objects to make extraction. The other framework is structured prediction which integrates the two subtasks into a unified structure and performs decoding in one step. \cite{r4} proposes a model using sequence tagging-based approaches and forbidding dependency trees. \cite{r5} employs graph convolutional networks for joint inference. \cite{r6} implements a table-filling strategy using a table encoder and a sequence encoder.

\subsection{Diffusion Model}

Diffusion model is a type of deep latent generative model primarily utilized for generating continuous data structures, such as images and audio. DDPM \cite{ddpm} is a pioneer work that makes diffusion model practical to applications, thus inviting excellent works \cite{d1, d2} in various fields. Recently, there has been an emergence of works in NLP utilizing diffusion models, such as \cite{d3, d4} in language model and \cite{d7,d5} in sequence-to-sequence tasks, despite the perceived challenges in applying diffusion models to discrete text sequences. Notably, DiffusionNER \cite{d6} also applies the diffusion model to named entity recognition. However, there are significant differences with our IPED, particularly in (1) task definition: IPED concentrates on extracting relational triples rather than mere entities. (2) core design: our model operates by diffusing in a three-dimensional space for each triple, in contrast to DiffusionNER, which diffuses within a one-dimensional matrix for each entity and incorporates an additional classifier.

\begin{figure*}[t]
  \vspace{-10pt}
  \centering
  \includegraphics[width=\textwidth]{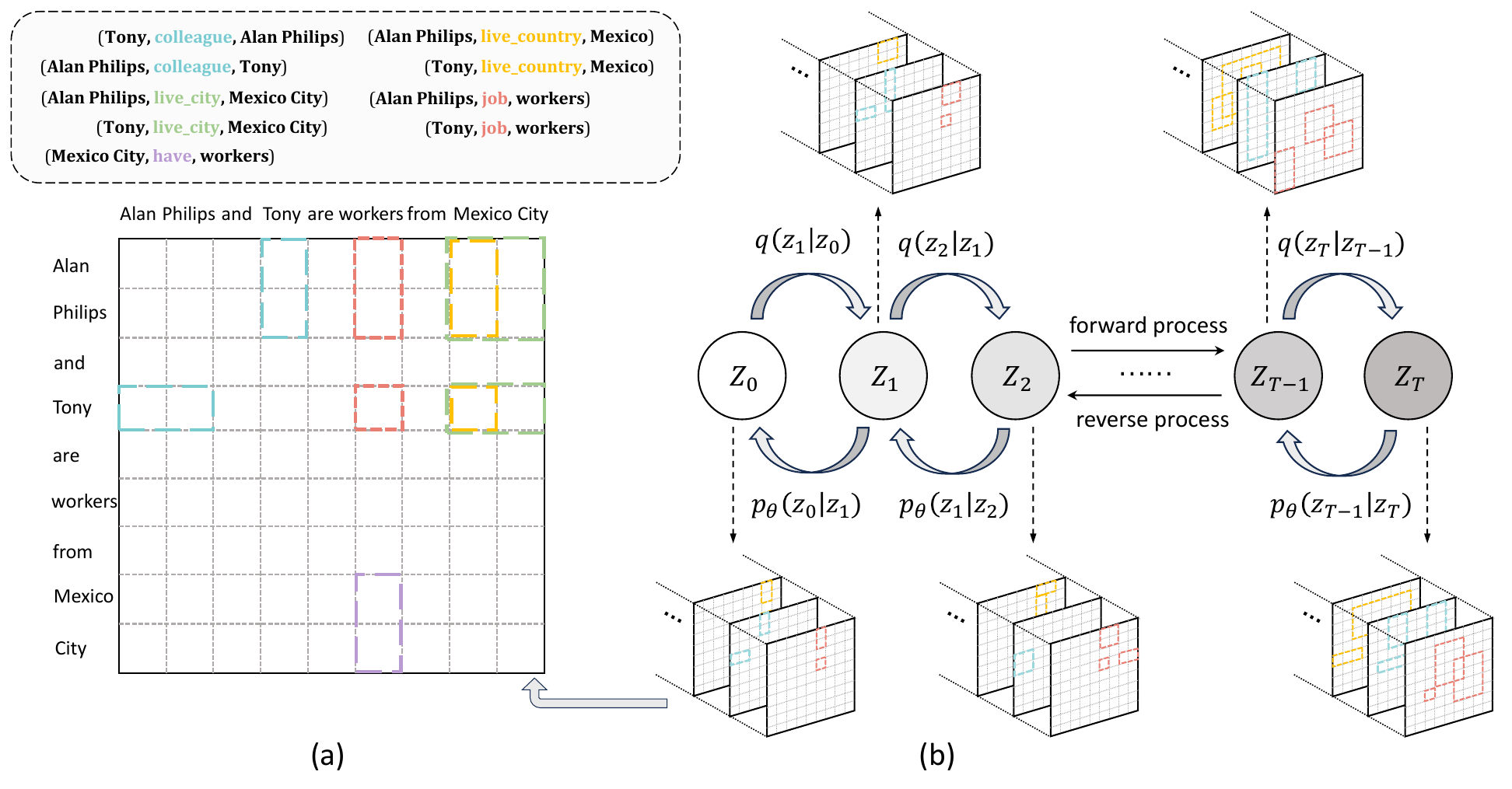}
  \caption{Figure (a) depicts our table-filling strategy along with triple demonstration. For the convenience of illustration, we simplify our three-dimensional tables (as in Figure (b)) into the form of a two-dimensional table in Figure (a), containing nine blocks in total that represent nine triples. Here, dashed rectangles denote the four edges of the blocks, and different colors indicate the levels of the blocks. Figure (b) illustrates the overall diffusion process.}
  \label{figure1}
\end{figure*}

\section{Methodology}

This section firstly introduces our implicit table-filling strategy and its corresponding decoding algorithm. Secondly, the formulation of the Block-Denoising Diffusion Model is presented. Finally, the network architecture of our model is detailed.

\subsection{Implicit Block-Covered Table Filling}

For a sentence \(\mathcal{S} = \left \{ x_{1}, x_{2},...,x_{L}  \right \}\) composed of L words, K relations \(\mathcal{R} = \left \{ r_{1}, r_{2},...,r_{K}  \right \}\) are predefined in a dataset. The objective of relational triple extraction is to identify all triples (\textit{head}, \textit{relation}, \textit{tail}) in each sentence, where the head and tail represent the subject and object entities, respectively, along with their connected relation. Within a sentence, for all triples \(\tau = \{ \left ( h_{i}, r_{i}, t_{i} \right )\}_{i=1}^{M} \), M denotes the total number of triples, and \(h_{i}, t_{i}\) represent the entity spans, each composed of one or more consecutive tokens.

Unlike previous classifier-based tagging methods, our model does not allocate a label to each unit of the L*L*K three-dimensional matrix. Instead, it refines M blocks (\(\mathbf{B}  \in \mathbb{R}^{M \times 5}\)) to cover the K tables horizontally, vertically, and deeply, which is, our implicit way to fill the tables. As illustrated in Figure \ref{figure1}, each block consists of five elements: the up and down edges indicate vertical positioning, the left and right edges denote horizontal positioning, and the level represents depth positioning within the K stacked tables, with each table corresponding to a specific relation. Via our proposed Blk-DDM (described in Section \ref{sec: Block-Denoising Diffusion Model}), these M blocks are progressively refined to reveal the recognized triples.

The proposed decoding scheme, named Parallel Boundary Emitting Strategy (PBES), is introduced to extract triples from the blocks. PBES follows the four edges and one level of each block, emitting them in parallel to the corresponding entities and relation. Specifically, for each block, the up and down edges are extended to the left side of the table, indicating the boundaries of the head entity. Similarly, the left and right edges are extended correspondingly to identify the boundaries of the tail entity. Meanwhile, the depth level where the block is located signifies a specific table, thereby indicating a particular relation. By repeating this process M times as described, all blocks are converted into relational triples.

Our table-filling method enables the precise extraction of all existing triples by circumventing the conflicts typically associated with explicit tagging. Thanks to the lack of inner constraints between the M blocks, this approach not only naturally tackles complex scenarios such as SEO, EPO, and SOO, but also overcomes issues like the failure of single-token entity extraction in \cite{onerel} and error association in \cite{grte,odrte}.

\subsection{Block-Denoising Diffusion Model}
\label{sec: Block-Denoising Diffusion Model}

In this section, we present the formulation of block generation as a denoising diffusion process and introduce our block-denoising diffusion model (Blk-DDM). As depicted in Figure \ref{figure1}, the diffusion model comprises a forward process that incrementally introduces noise to data samples and a reverse process that recovers the ground truth through step-by-step denoising. These two processes are synchronized to facilitate the learning of a network endowed with the denoising capability. During the inference phase, the diffusion model incrementally refines data samples through a multistep denoising process from a standard Gaussian distribution. Consequently, we convert our M blocks, composed of five elements (up, down, left, right, level), into index format \(\mathbf{B}  = \{\left ( u_{i}, d_{i}, l_{i}, r_{i}, v_{i} \right ) \}_{i=0}^{M}\) to support the denoising operations. Following \cite{ddpm}, the forward denoising process is simplified by computing \(\{\bar{\alpha } _{1},...,\bar{\alpha } _{T}\}\) from a predefined variance schedule \(\left \{ \beta _{t} \right \} _{t=0}^{T} \in \left ( 0,1 \right ) \), and thus noise injection in multiple steps can be integrated into one step as follows:

\begin{equation}
  q\left ( z_{t} \mid z_{0}  \right ) = \mathcal{N}\left ( z_{t}; \sqrt{\bar{\alpha}_{t}  }z_{0}, \left ( 1- \bar{\alpha}_{t} \right )\mathbf {I} \right )
  \label{eq1}
\end{equation}

where \(q\) represents the forward process from \(z_{0}\) to \(z_{t}\). \(z_{0}\) and \(z_{t}\) denote the original data and the noised data at timestep t, respectively. \(\mathbf {I}\) is the standard Gaussian distribution. Note that the fixed forward process depicted in Figure \ref{eq1} can be considered as a Markov chain.

\begin{figure*}[t]
  \vspace{-10pt}
  \centering
  \includegraphics[width=\textwidth]{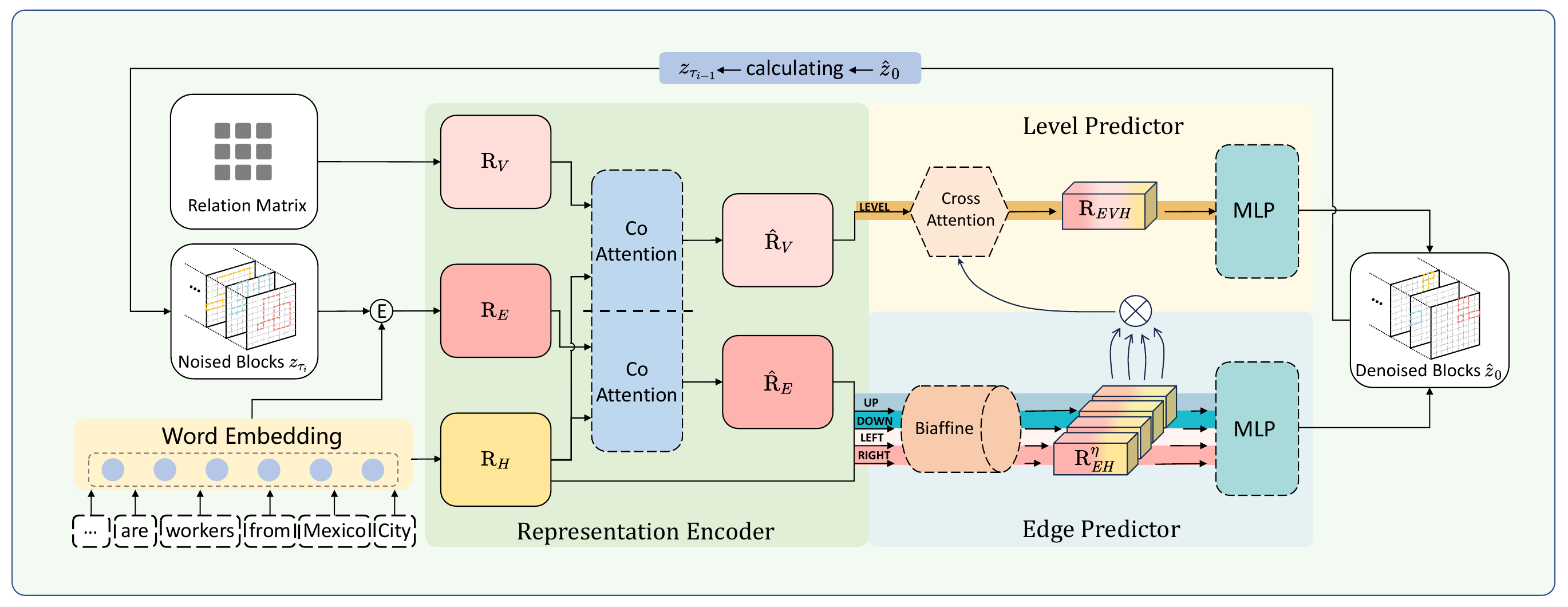}
  \caption{The overview model structure of IPED. To enhance the illustration of the workflow, we utilize three different colors to denote various feature representations: Pink for level information, Yellow for sentence information, and Red for edge information. \textcircled{\raisebox{-0.9pt}{\scriptsize E}} represents the encoding of \(\mathbf{R}_{\textit{E}}\). \(\otimes\) denotes the maxpooling operation. To simplify the illustration, the four Biaffine modules are integrated into one in this overview. To better display the reverse process as in Figure \ref{figure1}, a reverse-flow arrow is used to symbolize progressive denoising.
}
  \label{figure2}
\end{figure*}

\paragraph{Training Process}

The training process of the diffusion model involves a one-step noise addition and a one-step prediction towards the ground truth, aimed at training a network for inference purposes. As for a sentence, blocks \( \mathbf{B} \in \mathbb{R}^{M\times5}  \) are initially derived from M ground truth triples. Subsequently, \(\mathbf{B} \) is expanded by some blocks randomly sampled from a Gaussian distribution, resulting in \( z_{0} =\mathbf{B}  \in \mathbb{R}^{ N \times5} \) (N \(>\) M). Following Equation \eqref{eq1}, we then have

\begin{equation}
    z_{t} = \sqrt{\bar{\alpha}_{t}} z_{0} + \sqrt{1-\bar{\alpha}_{t} } \epsilon 
    \label{eq2}
\end{equation}

where t (\(\le \) predefined total timestep \(T\)) is a randomly chosen timestep and \(\epsilon\sim \mathcal{N}\left ( \textbf{0}, \mathbf {I} \right )\) donates the pure noise from the Gaussian distribution, thus getting noised blocks \(\mathbf{B} \). Feeding \(z_{t}\) into our network \(f_{\theta } \), one can get the predicted \(z_{0}\) (Section \ref{sec: Model Structure}) and compute the objective function (Section \ref{sec: Loss Function}). By optimizing the loss function, the weights of our network \(f_{\theta } \) will be updated accordingly.

\paragraph{Inference Process}

Following DDIM \cite{ddim}, the reverse diffusion process is defined as a non-Markovian chain to achieve inference acceleration. An arithmetic sequence \(\tau\) of length \(\sigma\) is predefined as \(\left [ 1,...,T \right ] \) and D purely noised blocks \(x_{T} \in \mathbb{R}^{D\times 5}  \) are sampled from the Gaussian distribution. Modified from DDIM, we have progressive denoising as follows:

\begin{flalign}
    & z_{\tau_{i-1}} = \sqrt{\bar{\alpha}_{\tau_{i-1}}} \hat{z}_{0} + \sqrt{1-\bar{\alpha}_{\tau_{i-1}}} \frac{z_{\tau_{i}} - \sqrt{\bar{\alpha}_{\tau_{i}}} \hat{z}_{0}}{\sqrt{1-\bar{\alpha}_{\tau_{i}}}} &
\end{flalign}

where \(\hat{z} _{0} \) is predicted by \(f_{\theta } \), with the index i traversing from \(\sigma\) to 1. After \(\sigma\) iterations, \(z_{0} \in \mathbb{R}^{D\times5}\) is recovered from the noise distribution. Note that D is a hyperparameter supposedly larger than the ground truth block number, and thus the filtration of predicted D blocks aims to minimize their divergence from the ground truth. Hence, blocks with the sum predicted probability below the threshold \(\varphi \) are discarded. \footnote{The probabilities, including \(\mathbf{P}^{\eta}\) and \(\mathbf{P}^{\textit{v}}\), will be explained in Section \ref{sec: Edge Predictor and Level Predictor}.}

\subsection{Model Structure}
\label{sec: Model Structure}
As shown in Figure \ref{figure2}, our model architecture consists of three parts: Representation Encoder, Edge Predictor, and Level Predictor. Accepting one sentence, noised blocks (with timestep t) as inputs, the model network \(f_{\theta } \) generates the predicted blocks \(\hat{z} _{0} \) appropriately.

\subsubsection{Representation Encoder}
Given an input sentence \(\mathcal{S} = \left \{ x_{1}, x_{2},...,x_{L}  \right \}\) composed of L words or indexes, here our sentence encoder consists of a pre-trained BERT \cite{bert} and a bi-directional LSTM \cite{lstm}. Utilizing our encoder, token embeddings along with positional embeddings as the input are transformed into contextualized sentence representation \( \mathbf{R}_{\textit{H}} \in \mathbb{R}^{L\times d} \). Then the inner span tokens are extracted from the word indexes indicated by the edges of our blocks, yielding the edge representation \(\mathbf{R}_{\textit{E}} \in \mathbb{R}^{N\times d}\) following mean-pooling. Differently, the level representation \(\mathbf{R}_{\textit{V}} \in \mathbb{R}^{N\times d}\) is derived directly from an embedding relation matrix \(\mathrm{R} \in \mathbb{R}^{K\times d}\), where each row represents a distinct relation type and K denotes the total number of predefined relation types. This matrix is regarded as a trainable parameter set in our model.

To better fuse both edge representation and level representation with contextualized information, we utilize the hierarchical Co-Attention mechanism in our model, which is proven to be effective with multimodal data \cite{co1}. Among the two Parallel Co-Attention modules in our model, we illustrate one of them as an example, which attends to the sentence representation \(\mathbf{R}_{\textit{H}}\) and the edge representation \(\mathbf{R}_{\textit{E}}\) simultaneously. An affinity matrix \(\mathbf{C} \in \mathbb{R}^{L\times N}\) that transforms sentence attention space into edge attention space, and the attention score vector \(\mathbf{a}^{\textit{e}} \in \mathbb{R}^{N}\) that optimizes the affinity, are calculated as follows:

\begin{equation}
    \mathbf{C} = \mathrm {tanh} \left ( \mathbf{R}_{\textit{H}}^{\textit{T}}\mathbf{W}_{\textit{b}}\mathbf{R}_{\textit{E}}\right )
    \label{eq3}
\end{equation}

\begin{equation}
    \mathbf{H}^{\textit{e}} = \mathrm {tanh}\left ( \mathbf{W}_{\textit{e}}\mathbf{R}_{\textit{E}} \,+ \,\left ( \mathbf{W}_{\textit{h}}\mathbf{R}_{\textit{E}} \right )\mathbf{C} \right ) 
    \label{eq4}
\end{equation}

\begin{equation}
    \mathbf{a}^{\textit{e}} = \mathrm {softmax}\left ( \mathbf{w}_{\textit{he}}^{\textit{T}}\mathbf{H}^{\textit{e}} \right ) 
    \label{eq5}
\end{equation}
\vspace{2pt}

where \(\mathbf{W}_{\textit{b}} \in \mathbb{R}^{d\times d}\), \(\mathbf{W}_{\textit{e}} \in \mathbb{R}^{k\times d}\), \(\mathbf{W}_{\textit{h}} \in \mathbb{R}^{k\times d}\), \(\mathbf{w}_{\textit{he}} \in \mathbb{R}^{k}\) are learnable parameters, \(\mathbf{H}^{\textit{e}}\) is the middle state. Finally, the edge attention vector \(\hat{\mathbf{R}}_{\textit{E}} \in \mathbb{R}^{N\times d}\) is calculated as the weighted sum of the edge features plus an additional sinusoidal embedding \cite{attention}:

\begin{equation}
    \hat{\mathbf{R}}_{\textit{E}} = \mathbf{a}^{\textit{e}}\mathbf{R}_{\textit{E}} + \mathbf{E}_{\textit{t}}
    \label{eq6}
\end{equation}

where \( \mathbf{E}_{\textit{t}}\) is the embedding of timestep t. Equally, the same operation is implemented to obtain the fused level representation \(\hat{\mathbf{R}}_{\textit{V}} \in \mathbb{R}^{N\times d}\).

\begin{table*}[ht]
\centering
\vspace{-10pt}
\renewcommand{\arraystretch}{1.4}
\small
\begin{tabular}{llllcccccccccccc}
\toprule
\multicolumn{4}{c}{\multirow{2}{*}{Method}} & \multicolumn{3}{c}{NYT*} & \multicolumn{3}{c}{WebNLG*} & \multicolumn{3}{c}{NYT} & \multicolumn{3}{c}{WebNLG} \\ \cmidrule(lr){5-7} \cmidrule(lr){8-10} \cmidrule(lr){11-13} \cmidrule(lr){14-16}
\multicolumn{4}{c}{}                        & Prec.   & Rec.   & F1    & Prec.    & Rec.    & F1     & Prec.   & Rec.  & F1    & Prec.    & Rec.   & F1     \\ \hline
\multicolumn{4}{l}{GraphRel \cite{graphrel}}                & 63.9    & 60.0   & 61.9  & 44.7     & 44.1    & 42.9   & -       & -     & -     & -        & -      & -      \\
\multicolumn{4}{l}{RSAN \cite{rsan}}                     & -    & -   & -  & -     & -    & -   & 85.7    & 83.6  & 84.6  & 80.5        & 83.8      & 82.1      \\
\multicolumn{4}{l}{TPLinker \cite{tplinker}}                & 91.3    & 92.5   & 91.9  & 91.8     & 92.0    & 91.9   & 91.4    & 92.6  & 92.0  & 88.9     & 84.5   & 86.7   \\
\multicolumn{4}{l}{GRTE \cite{grte}}                    & 92.9    & 93.1   & 93.0  & 93.7     & 94.2    & 93.9   & 93.4    & 93.5  & 93.4  & 92.3     & 87.9   & 90.0   \\
\multicolumn{4}{l}{PRGC \cite{prgc}}                    & 93.3    & 91.9   & 92.6  & 94.0     & 92.1    & 93.0   & 93.5    & 91.9  & 92.7  & 89.9     & 87.2   & 88.5   \\
\multicolumn{4}{l}{EmRel \cite{emrel}}                   & 91.7    & 92.5   & 92.1  & 92.7     & 93.0    & 92.9   & 92.6    & 92.7  & 92.6  & 90.2     & 87.4   & 88.7   \\
\multicolumn{4}{l}{RelU-Net \cite{relu-net}}                & 93.3    & 92.9   & 93.1  & 94.9     & 93.7    & 94.3   & -       & -     & -     & -        & -      & -      \\
\multicolumn{4}{l}{BiRTE \cite{birte}}                   & 92.2    & 93.8   & 93.0  & 93.2     & 94.0    & 93.6   & 91.9    & \textbf{93.7}  & 92.8  & 89.0     & 89.5   & 89.3   \\
\multicolumn{4}{l}{OneRel \cite{onerel}}                  & 92.8    & 92.9   & 92.8  & 94.1     & 94.4    & 94.3   & 93.2    & 92.6  & 92.9  & 91.8     & 90.3   & 91.0   \\
\multicolumn{4}{l}{RFBFN \cite{rfbfn}}                   & 93.4    & 93.2   & 93.3  & 93.9     & 94.1    & 94.0   & 93.7    & 93.6  & 93.6  & 91.5     & 89.4   & 90.4   \\
\multicolumn{4}{l}{ODRTE \cite{odrte}}                   & 93.5    & \textbf{93.9}   & 93.7  & 94.6     & 95.1    & 94.9   & 94.2    & 93.6  & 93.9  & 92.8     & 92.1   & 92.5   \\ \hline
\multicolumn{4}{l}{IPED}                    & \textbf{94.2}      & 93.5     & \textbf{93.9}    & \textbf{95.3}       & \textbf{95.7}      & \textbf{95.5}     & \textbf{94.7}      & 93.4    & \textbf{94.1}    & \textbf{93.0}       & \textbf{93.6}     & \textbf{93.3}        \\ \bottomrule
\end{tabular}
\caption{Main results of IPED and other baselines.}
\label{table1} 
\end{table*}

\subsubsection{Edge Predictor and Level Predictor}
\label{sec: Edge Predictor and Level Predictor}

For the Edge Predictor, we employ Biaffine to acquire fine-grained fused representations, which is proposed for dependency parsing \cite{biaffine} at the outset. Here we have four Biaffine for \(\mathbf{R}_{\textit{EH}} ^{\eta}\) representations where \(\eta \in \left \{ u,d,l,r \right \} \) symbolizes four edges, respectively. \(\mathbf{R}_{\textit{EH}} ^{\eta}\) is obtained as follows:

\begin{align}
    \mathbf{R}_{\textit{EH}} ^{\eta} &= \mathrm{B\hspace{0.5pt}i\hspace{0.5pt}a\hspace{0.5pt}f\hspace{0.5pt}f}^{\eta}\left ( \mathbf{R}_{\textit{H}}, \hat{\mathbf{R}}_{\textit{E}} \right ) \nonumber \\
                   &= \mathbf{R}_{\textit{H}}^{\textit{T}}\mathbf{U}_{1}^{\eta}\hat{\mathbf{R}}_{\textit{E}} + \mathbf{U}_{2}^{\eta}\left ( \mathbf{R}_{\textit{H}} \oplus \hat{\mathbf{R}}_{\textit{E}} \right ) +  \mathbf{b}^{\eta}
    \label{eq7}
\end{align}

where \(\mathbf{U}_{1}^{\eta}\) and \(\mathbf{U}_{2}^{\eta}\) donate two parameter matrices, \(\mathbf{b}^{\eta}\) is the bias vector, \(\oplus\) means concatenation. Then \(\mathbf{R}_{\textit{EH}} ^{\eta}\) are put through four simple multiple-layer perceptrons with softmax layers to get the probabilities \(\mathbf{P}^{\eta} \in \mathbb{R}^{N\times L}\) for four edges in blocks.

For the Level Predictor, a cross-attention layer is utilized to obtain the deep latent representation \(\mathbf{R}_{\textit{EVH}}\), incorporating edge-sentence embedding \(\mathbf{R}_{\textit{EH}} ^{\eta}\) to level representation \(\hat{\mathbf{R}}_{\textit{V}}\). Specifically, \(\mathbf{R}_{\textit{EH}} ^{\eta}\) undergoes a max-pooling operation to serve as the key and value tensors, while \(\hat{\mathbf{R}}_{\textit{V}}\) acts as the query tensor. Then the level probability \(\mathbf{P}^{\textit{v}} \in \mathbb{R}^{N\times K}\) is determined using a multilayer perceptron, followed by a softmax layer.

\subsubsection{Loss Function}
\label{sec: Loss Function}

In conjunction with the predicted probabilities above, the Log-Likelihood Function is maximized to train our model parameters. As N blocks are generated during training, yet only M ground truth blocks exist, we solve the optimal match via the Hopcroft-Krap algorithm \cite{HK}. Our objective function is defined as follows:

\begin{equation}
\begin{split}
\mathcal{L} = & -\sum_{i=1}^{N} \biggl[ \beta_{1} \sum_{\eta \in \{ u,d \}} \log \mathbf{P}_{i}^{\eta} \bigl( \xi^{\eta}(i) \bigr) \\
&\quad + \beta_{2} \sum_{\eta \in \{ l,r \}} \log \mathbf{P}_{i}^{\eta} \bigl( \xi^{\eta}(i) \bigr) \\
&\quad + \beta_{3} \log \mathbf{P}_{i}^{\textit{v}} \bigl( \xi^{\textit{v}}(i) \bigr) \biggr]
\end{split}
\end{equation}

where \(\xi\left( i \right )\) represents the ground truth edges and level of the \(i\)-th block, \(\beta_{1}, \beta_{2}, \beta_{3}\) are the hyperparameters for the weights of each prediction part.

\section{Experiments}

\subsection{Datasets}

Following previous works \cite{onerel,odrte}, we evaluate our model on two well-known datasets NYT \cite{nyt} and WebNLG \cite{webnlg}. The NYT dataset is extracted using the distantly supervised method from New York Times news articles, while the WebNLG dataset was originally designed for Natural Language Generation. Each dataset exists in two versions: one is annotated with the whole entity span, and the other is annotated with the last word of entities. For clarity, we mark the fully annotated version as NYT and WebNLG, and the simpler annotated version as NYT* and WebNLG*, respectively. Following prior works, we split the test set of each dataset based on the number of triples and the overlapping pattern in each sentence.

\begin{table*}[ht]
\centering
\setlength{\tabcolsep}{4pt} 
\vspace{-10pt}
\renewcommand{\arraystretch}{1.4}
\small
\begin{tabular}{llllcccccccccccccccc}
\toprule
\multicolumn{4}{c}{\multirow{2}{*}{Model}} & \multicolumn{8}{c}{NYT*} & \multicolumn{8}{c}{WebNLG*} \\ \cmidrule(lr){5-12} \cmidrule(lr){13-20}
\multicolumn{4}{c}{} & Normal & SEO & EPO & Q=1 & Q=2 & Q=3 & Q=4 & Q\(\ge\)5 & Normal & SEO & EPO & Q=1 & Q=2 & Q=3 & Q=4 & Q\(\ge\)5 \\ 
\midrule
\multicolumn{4}{l}{GRTE} & 91.1 & 94.4 & 95.0 & 90.8 & 93.7 & 94.4 & 96.2 & 93.4 & 90.6 & 94.5 & 96.0 & 90.6 & 92.5 & 96.5 & 95.5 & 94.4 \\
\multicolumn{4}{l}{PRGC} & 91.0 & 94.0 & 94.5 & 91.1 & 93.0 & 93.5 & 95.5 & 93.0 & 90.4 & 93.6 & 95.9 & 89.9 & 91.6 & 95.0 & 94.8 & 92.8 \\
\multicolumn{4}{l}{RFBFN} & 91.2 & 95.2 & 95.6 & 91.4 & \textbf{93.8} & 94.8 & 96.4 & 93.9 & 91.0 & 94.6 & 96.5 & 90.8 & 92.6 & 96.6 & 94.7 & 94.5 \\
\multicolumn{4}{l}{ODRTE} & \textbf{91.3} & 95.7 & 95.9 & 91.3 & 93.4 & 94.6 & 96.9 & 95.3 & 92.1 & 95.4 & 95.9 & 91.1 & 93.5 & 95.9 & 96.1 & 95.1 \\ 
\midrule
\multicolumn{4}{l}{IPED} & 91.0 & \textbf{95.7} & \textbf{96.0} & \textbf{91.5} & 93.2 & \textbf{94.9} & \textbf{97.3} & \textbf{95.4} & \textbf{92.1} & \textbf{95.6} & \textbf{96.9} & \textbf{91.8} & \textbf{94.2} & \textbf{96.8} & \textbf{96.7} & \textbf{96.0} \\ 
\bottomrule
\end{tabular}
\caption{F1 score on sentences with different overlapping patterns and different triple numbers. Q stands for the number of triples in a sentence.}
\label{table2}
\end{table*}

\subsection{Evaluation Metrics}

For a fair comparison with prior works mentioned above, we report standard micro Precision (Prec.), Recall (Rec.), and F1-score (F1.) as our three evaluation metrics. Meanwhile, we implement distinct matching rules for each version of the datasets. In the case of NYT and WebNLG datasets, an extracted relational triple is regarded correct only if all words of both entities and the relation type precisely align with the ground truth. For NYT* and WebNLG* datasets, only the last words of two entities and the relation are required to be correct.

\subsection{Implementation Details}

To make a fair comparison, we utilize the cased base version of BERT \cite{bert} as our pretrained model. The AdamW optimizer \cite{adamw} is employed with a learning rate of 3\(e\)-5. The hidden size of our cross-attention and biaffine modules is configured to 1024. A warm-up learning rate scheduler, with a 0.1 ratio and a maximum gradient normalization of 1.5, is configured for the training process. Regarding the diffusion setting, the total timestep T is set to 1000, the sampling timestep \(\sigma\) to 10, and the number of denoising blocks D to 30. The sum threshold \(\varphi \) for the edges and level probabilities is established at 4.

\subsection{Main Results}

Table \ref{table1} presents the performance comparison between our IPED and various baselines across four benchmarks. It can be seen that our model, IPED, outperforms all the baselines and achieves state-of-the-art performance, even when compared to the strongest explicit table-filling baseline ODRTE \cite{odrte} and the leading multi-task joint framework RFBFN \cite{rfbfn}. This proves the dramatic efficacy of our implicit perspective and denoising diffusion strategy.

Compared with the best baseline ODRTE, our IPED achieves a 0.2 absolute improvement in F1-score on both NYT and NYT*. It is worth noticing that, a significant improvement, 0.8 and 0.6 gains in F1-score, is achieved on WebNLG and WebNLG* respectively, whereas many models \cite{tplinker, ergm} blame their poor performance on the complexity arising from hundreds of predefined relation types. We attribute our advancement on large-relation datasets to block-level progressive refinement; specifically, our block-denoising diffusion model allows fine-tuned block denoising across various levels of the tables.

The results on NYT and WebNLG reveal that our IPED outperforms OneRel \cite{onerel} by 1.2\% and 2.3\%, and GRTE \cite{grte} by 0.7\% and 3.3\% in terms of F1-score, respectively. This demonstrates that the implicit table-filling scheme can immensely avoid interruptions caused by redundant negative tagging, which otherwise leads to negative bias. This improvement highlights two key advantages of our approach: the capability to recognize all potential triples and the proficiency in avoiding error association during decoding.

\subsection{Performance on Complex Scenarios}

To validate the ability of our model to handle diverse overlapping patterns and multiple triples, we conduct further experiments on NYT* and WebNLG*. As indicated in Table \ref{table2}, our proposed IPED model surpasses nearly all baselines on both datasets, with the exception of two scenarios on NYT* when Q equals 2 and when there is no overlap. In complex scenarios, such as multiple triples within a single sentence, the performance of IPED turns out to be exceptional, surpassing four state-of-the-art models. The reason behind this is that our decoding scheme, the Parallel Boundary Emitting Strategy (PBES), has the capacity to accurately map our blocks into ground truth triples. This contrasts with previous decoding algorithms in explicit table-filling methods \cite{grte}, which often incorrectly decode triples due to error association.

\begin{table*}[ht]
\centering
\vspace{-10pt}
\setlength{\tabcolsep}{4pt} 
\renewcommand{\arraystretch}{1.4}
\small
\begin{tabular}{lcccccccc}
\toprule
\multirow{2}{*}{Model} & \multicolumn{4}{c}{NYT} & \multicolumn{4}{c}{WebNLG} \\
\cmidrule(lr){2-5} \cmidrule(lr){6-9}
 & Training Time & GPU Mem & Infer. Time (1/8) & F1 & Training Time & GPU Mem & Infer. Time (1/8) & F1 \\ 
\midrule
GRTE & \(931^{\dagger } \) & \(18771^{\dagger } \) & 44.1 / 9.6 & 93.4 & \(118^{\dagger } \) & \(15345^{\dagger } \) & 62.4 / 15.6 & 90.0 \\
OD-RTE & \(\textbf{798}^{\dagger } \) & \(8372^{\dagger } \) & 38.3 / 8.4 & 93.9 & \(\textbf{70}^{\dagger } \) & \(7515^{\dagger } \) & 51.0 / 12.8 & 92.5 \\ 
\midrule
\(\mathrm {IPED }_{\left [ \sigma =5 \right ] } \) & 887 & 5636 & \textbf{22.1} / \textbf{4.7} & 94.0 & 102 & 3778 & \textbf{30.1} / \textbf{7.7} & 93.1 \\
\(\mathrm {IPED }_{\left [ \sigma =10 \right ] } \) & 887 & 5636 & 26.6 / 5.8 & 94.1 & 102 & 3778 & 35.5 / 8.7 & 93.3 \\
\(\mathrm {IPED }_{\left [ \sigma =15 \right ] } \) & 887 & \textbf{5636} & 33.4 / 7.2 & \textbf{94.2} & 102 & \textbf{3778} & 40.6 / 10.2 & \textbf{93.4} \\ 
\bottomrule
\end{tabular}
\caption{Comparison of model efficiency. Training Time means the time (seconds) to train one epoch. GPU Mem stands for memory (MB) occupation during inference with the batch size of 8, and Infer. Time (1/8) donates the time (ms) to process each sentence with the batch sizes of 1 and 8, respectively. The superscript \(\dagger\) indicates the results reported by OD-RTE. All experiments are conducted on a single GeForce RTX 3090 with default configuration.}
\label{table3}
\end{table*}

\subsection{Computational Efficiency}
\label{sec: computational efficiency}

To evaluate the computational efficiency of our IPED, we conduct further experiments with respect to \textit{Training Time}, \textit{GPU Memory}, \textit{Inference Time}, and \textit{F1-score} on NYT and WebNLG. As demonstrated in Table \ref{table3}, we selected two robust baselines, GRTE and OD-RTE, for comparison. To verify the impact of the sampling timestep, we execute IPED with varying \(\tau\) values. It can be seen that when \(\sigma=5\), the inference speed of IPED is more than double that of GRTE, and it requires the least GPU memory compared to both baselines. Due to the inherent nature of diffusion training, the training time of our model is not the shortest, falling between OD-RTE and GRTE. Nevertheless, our IPED achieves a superior F1-score and greater inference efficiency. We conjecture the reasons might be our implicit table-filling strategy, which is exempt from redundant tagging, and the non-Markovian process employed during 
sampling.

\begin{figure}[!htbp]
  \centering
  \includegraphics[width=\linewidth]{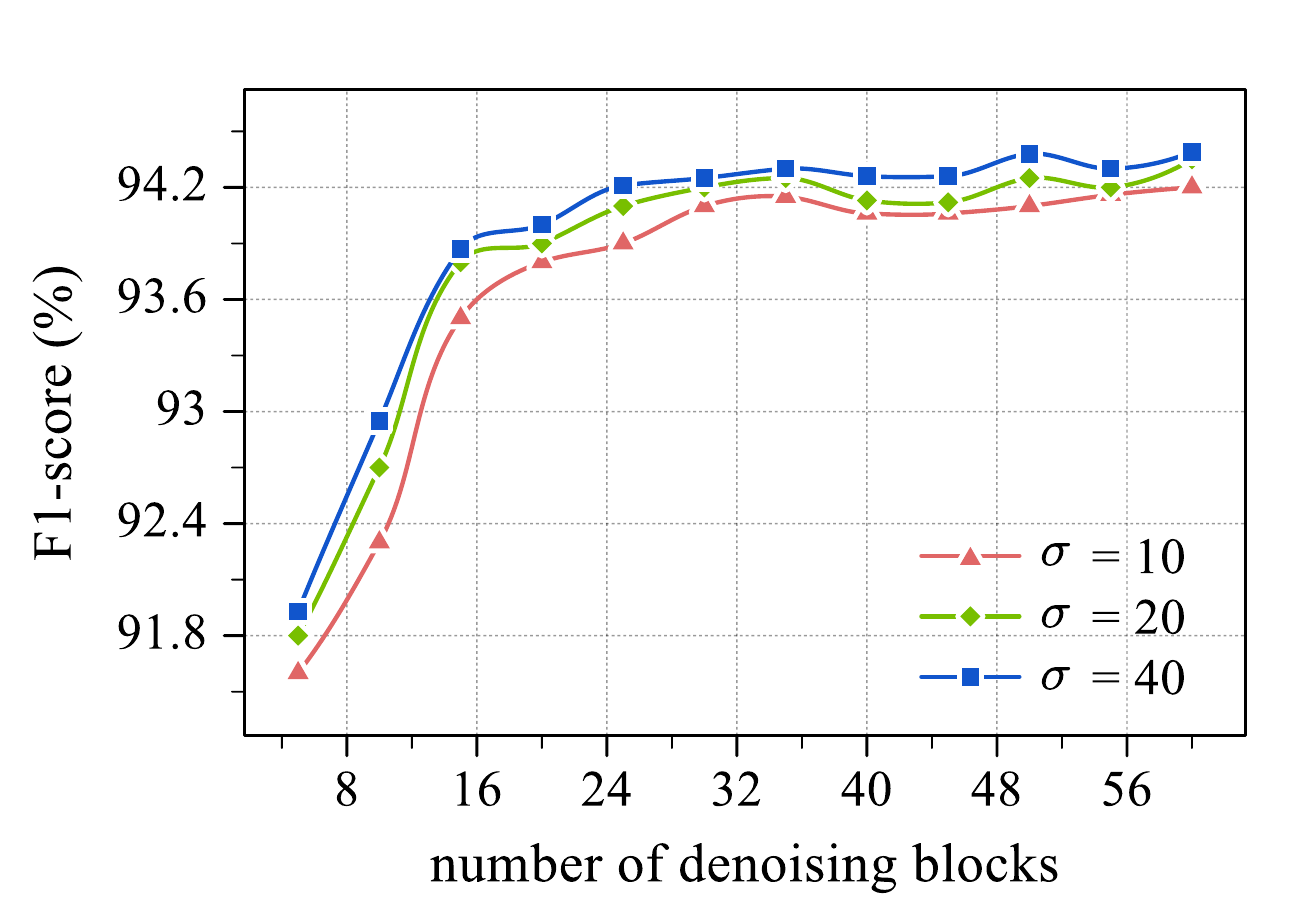}
  \caption{Performance of IPED with different number of denoising blocks D in terms of F1-score on NYT.}
  \label{figure3}
\end{figure}

\begin{figure}[!htbp]
  \centering
  \includegraphics[width=\linewidth]{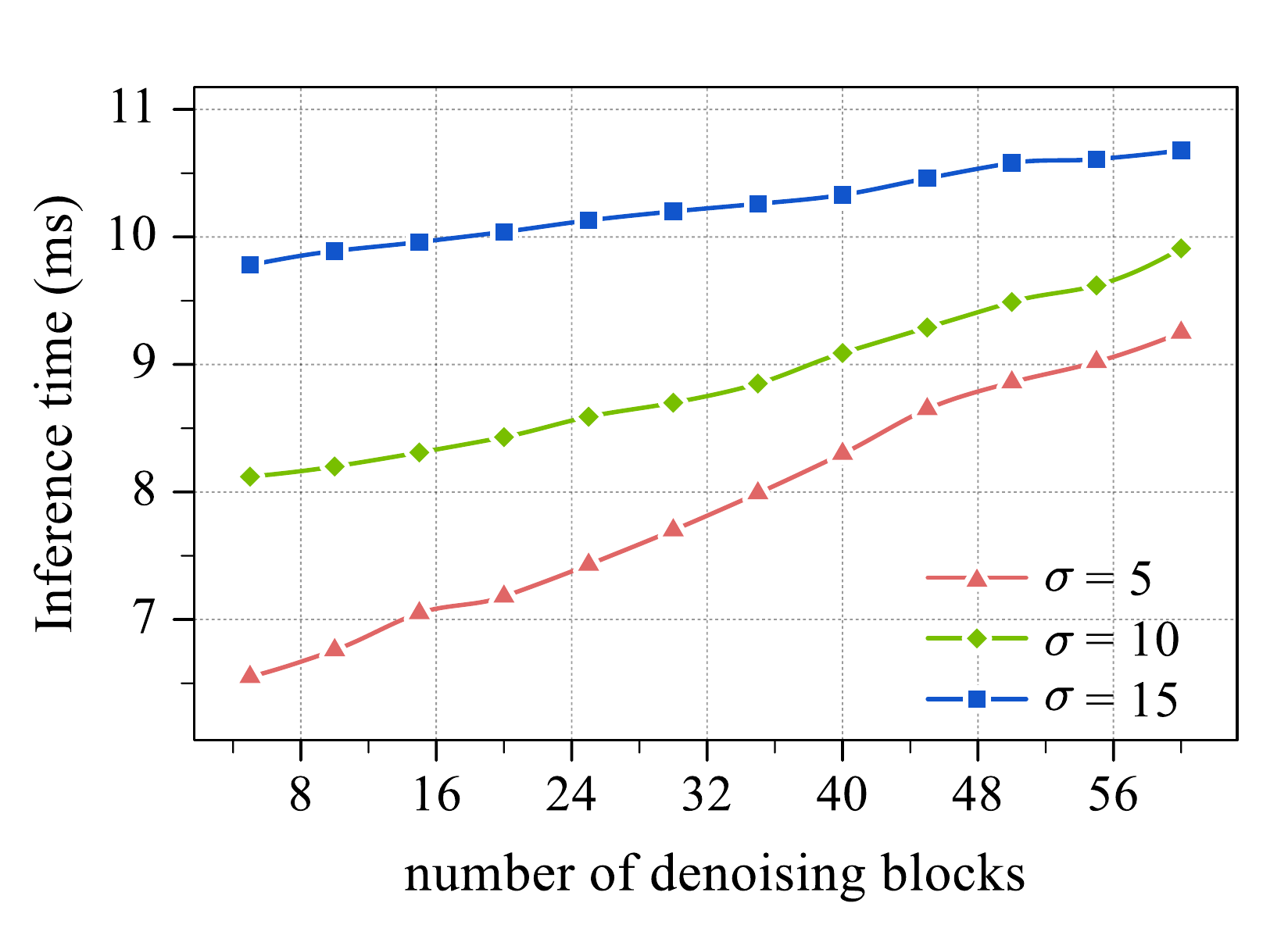}
  \caption{Performance of IPED with different number of denoising blocks D in terms of inference time on WebNLG. Note that the batch size is 8 during inference.}
  \label{figure4}
\end{figure}

\subsection{Analysis on Sampling Number}

In the denoising inference process, the number of denoising blocks, denoted as D, is a crucial parameter. We conducted additional experiments on it with different sampling timestep \(\sigma\) to evaluate its impact on F1-score and inference time. As depicted in Figure \ref{figure3}, the F1-score decreases sharply when D is less than 15 and remains stable when D exceeds 25. It can be observed from Figure \ref{figure4} that the inference time increases with larger D values, especially when \(\sigma\) is relatively small. Regarding the sampling timestep \(\sigma\), these two figures indicate that a larger \(\sigma\) brings about a higher F1-score but also increases inference time. To balance the F1-score and inference time, we set D at 30 and \(\sigma\) at 10 as our standard configuration. Consequently, our IPED is capable of properly covering all potential blocks, thereby enhancing the recall rate while ensuring optimal inference time for practical applications.

\begin{table}[t]
\vspace{30pt}
\centering
\begin{tabular}{lllccc}
\toprule
\multicolumn{3}{l}{Model} & P & R & F \\ 
\midrule
\multicolumn{3}{l}{IPED} & \textbf{93.0} & \textbf{93.6} & \textbf{93.3} \\ 
\cdashline{1-6}
\multicolumn{3}{l}{\hspace{0.6em}w/o Co-Attention} & 91.9 & 92.2 & 92.1 \\
\multicolumn{3}{l}{\hspace{0.6em}w/o Biaffine} & 92.2 & 93.0 & 92.6 \\
\multicolumn{3}{l}{\hspace{0.6em}w/o Cross Attention} & 92.1 & 92.5 & 92.3 \\ 
\cdashline{1-6}
\multicolumn{3}{l}{\hspace{0.6em}w/o Level} & 90.6 & 91.6 & 91.1 \\ 
\bottomrule
\end{tabular}
\caption{Ablation study on WebNLG dataset.}
\label{table4}
\end{table}

\subsection{Ablation Study}
\label{Ablation}

Ablation experiments are conducted to explore the contributions of the primary components within the network architecture and the effectiveness of level diffusion, as shown in Table \ref{table4}. Observations reveal that removing any of the three components leads to a relative performance drop. Each of these three components is a critical part for representation construction, with the Co-Attention module having the most influence. Upon replacing the Co-Attention module with the simple addition of two input representations, a 1.2\% F1 decline is observed. The experiments indicate that all three modules in our network play a crucial role in recovering blocks from noise.

It is noteworthy that the performance decreases by 2.2\% when Level is omitted. This implies that IPED abandons the denoising diffusion process at the block Level, transitioning the task from three-dimensional to two-dimensional denoising. Specifically, noisy blocks are distributed across each level of the three-dimensional tables, with each block constrained to denoising at a specific level, thus precluding the possibility of progressive refinement with the block level. Thus it can be concluded that block-level denoising is crucial for the effectiveness of our block-denoising diffusion model in identifying triple relations, particularly in large-relation datasets like WebNLG.

\section{Conclusion}

This paper proposes an implicit approach to relational triple extraction, diverging from the explicit tagging methods of prior table-filling methods, thereby addressing several prevailing issues. Via denoising the edges and levels of noisy blocks, our introduced block-denoising diffusion model incrementally generates ground truth blocks, which can be swiftly and precisely converted into triples with our decoding algorithm PBES. Moreover, our network architecture incorporates beneficial modules such as Co-Attention and Biaffine, which promote the fusion of diverse representations. Experimental results on public datasets demonstrate that our IPED exceeds the performance of state-of-the-art (SoTA) models, while also achieving significantly faster inference speeds.

\section*{Limitations}

Two limitations of IPED warrant discussion. Firstly, IPED exhibits a substantial increase in training time consumption compared to some models, as detailed in Section \ref{sec: computational efficiency}. This can be attributed to the extensive denoising timestep required for training, leading to slow and fluctuating convergence, thereby necessitating a greater number of training epochs. Secondly, the application of our implicit perspective is currently limited to relational triple extraction. Such perception holds potential for broader application in information extraction tasks such as document-level relation extraction and event extraction, addressing the issue of redundant negative tagging inherent in table-filling. These possibilities will be explored in our future work.

\section*{Acknowledgements}

The authors wish to thank all anonymous reviewers for their valuable comments. This work was supported by the Shenzhen Science and Technology Program (JCYJ20230807094104009).

\bibliography{acl_latex}

\vspace{30pt}

\appendix

\begin{table*}[ht]
\centering
\setlength{\tabcolsep}{4pt} 
\renewcommand{\arraystretch}{1.4}
\small
\begin{tabular}{ccccccccccccc}
\toprule
\multirow{2}{*}{Dataset} & \multicolumn{3}{c}{Sentences} & \multicolumn{9}{c}{Details of test set} \\ \cmidrule(lr){2-4} \cmidrule(lr){5-13}
 & Train & Valid & Test & Normal & SEO & EPO & SOO & Q=1 & Q=2 & Q\textgreater{}2 & Relations & Triples \\ 
\midrule
NYT & 56196 & 5000 & 5000 & 3071 & 1273 & 1168 & 117 & 3089 & 1047 & 864 & 24 & 8616 \\
NYT* & 56195 & 4999 & 5000 & 3266 & 1297 & 978 & 45 & 3244 & 1045 & 711 & 24 & 8110 \\
WebNLG & 5019 & 500 & 703 & 239 & 448 & 6 & 85 & 256 & 175 & 272 & 216 & 1607 \\
WebNLG* & 5019 & 500 & 703 & 245 & 457 & 26 & 84 & 266 & 171 & 266 & 171 & 1591 \\
\bottomrule
\end{tabular}
\caption{Statistics of datasets used in our experiments. Q represents the number of triples in a sentence. Note that a single sentence can simultaneously contain SEO, EPO and SOO overlapping patterns.}
\label{table5}
\end{table*}

\section{Dataset Statistics}

The statistical details of the two datasets are displayed in Table \ref{table5}.

\section{Clarification for D, N and M}

In our paper, N is the number of blocks after expansion for training, M is the number of ground truth blocks for training, and D is the number of initialized blocks for inference. 

During training, there are M blocks at first, which are then expanded by adding N-M randomly sampled blocks, resulting in a total of N blocks. Specifically, N and D are two similar hyperparameters; N is used for training while D is for inference, and typically, both are larger than M. To clearly distinguish between training and inference in our paper, we have defined N and D separately for the readers.

\end{document}